\documentclass{article} 
\usepackage{acra}
\usepackage{cite}
\usepackage{tikz}
\usepackage{amsmath,amssymb,amsfonts}
\usepackage{algorithmic}
\usepackage{graphicx}
\usepackage{subcaption}
\graphicspath{ {figs/} }
\usepackage{textcomp}
\usepackage{float}
\usepackage{xcolor}
\usepackage{booktabs}
\usepackage{natbib}
\usepackage{flushend}
\def\BibTeX{{\rm B\kern-.05em{\sc i\kern-.025em b}\kern-.08em
    T\kern-.1667em\lower.7ex\hbox{E}\kern-.125emX}}

\title{\LARGE \bf Sim-to-Real Transfer of Robot Learning with Variable Length Inputs}
\author{Vibhavari Dasagi, Robert Lee, Serena Mou, Jake Bruce, Niko S\"underhauf, J\"urgen Leitner \\ Queensland University of Technology (QUT), Brisbane, Australia\\ Contact: vibhavari.dasagi@hdr.qut.edu.au}
\begin{document}

\thispagestyle{empty}
\pagestyle{empty}

\maketitle

\begin{abstract}

Current end-to-end deep Reinforcement Learning (RL) approaches require jointly learning perception, decision-making and low-level control from very sparse reward signals and high-dimensional inputs, with little capability of incorporating prior knowledge. This results in prohibitively long training times for use on real-world robotic tasks. Existing algorithms capable of extracting task-level representations from high-dimensional inputs, e.g. object detection, often produce outputs of varying lengths, restricting their use in RL methods due to the need for neural networks to have fixed length inputs.
In this work, we propose a framework that combines \textit{deep sets} encoding, which allows for variable-length abstract representations, with modular RL that utilizes these representations, decoupling high-level decision making from low-level control.  
We successfully demonstrate our approach on the robot manipulation task of object sorting,
showing that this method can learn effective policies within mere minutes of highly simplified simulation. The learned policies can be directly deployed on a robot without further training, and generalize to variations of the task unseen during training.

\end{abstract}

\section{Introduction}

Reinforcement Learning (RL) approaches for robotic control are an exciting research direction that shows promise in enabling robots to develop new skills autonomously.
However, while deep RL has seen success in a wide variety of simulated domains, there are still many limitations to be overcome before robotic systems can learn directly from real-world experience within a reasonable time frame. Training on a robot is costly in terms of both time and mechanical stress, and
perception and primitive motor control are amongst the most challenging aspects of learning behavior end-to-end. This is due largely to the richness and high-dimensionality of the data and dynamics involved, both of which are difficult to simulate. A huge amount of training experience is required for policies to succeed at such problems, resulting in a prohibitively large amount of time taken for robots to learn new skills~\citep{kalashnikov2018qt}.

\begin{figure}[t!]
	\centering
	\includegraphics[width=0.75\linewidth]{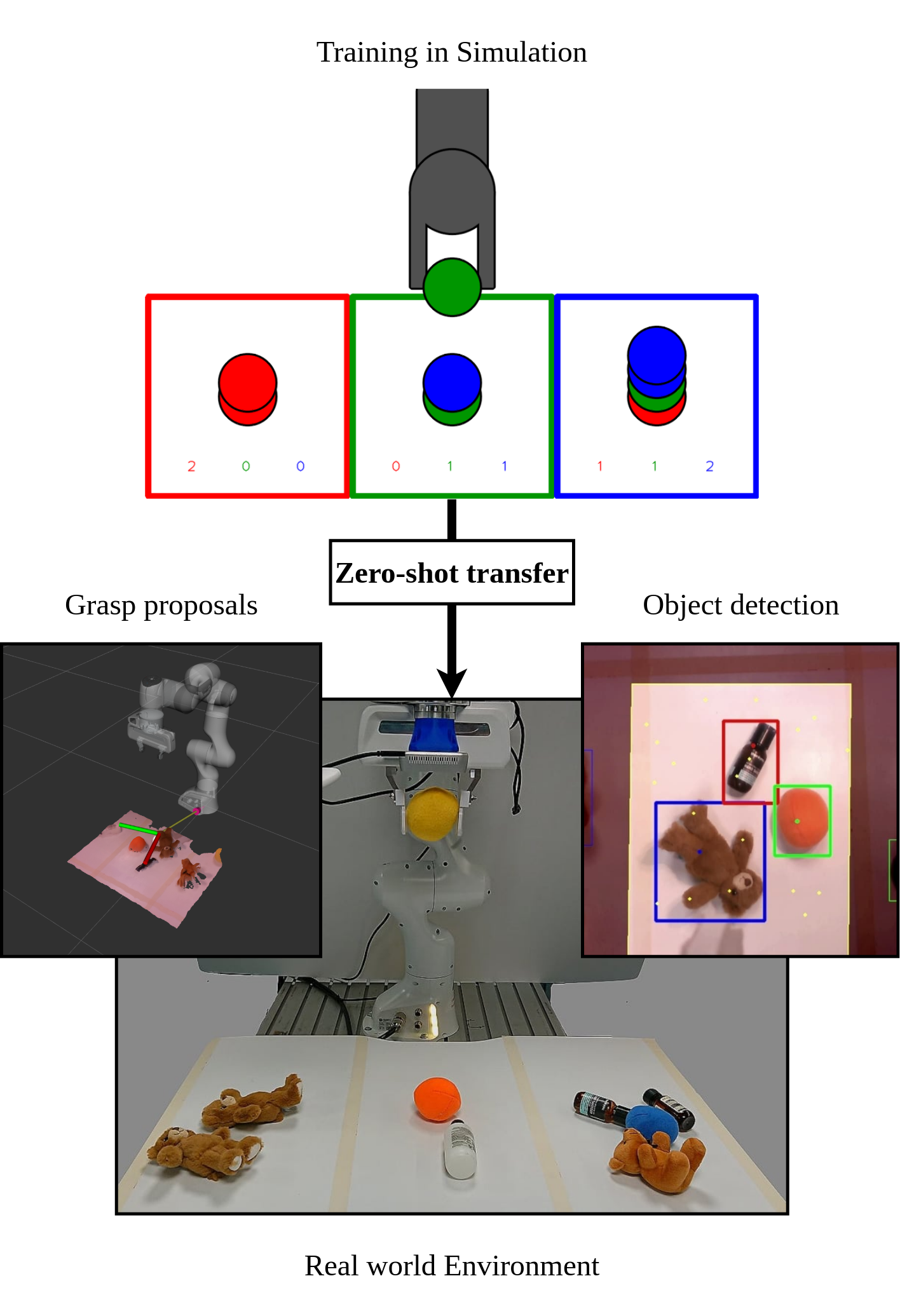}
	\caption{\small Sim-to-real transfer of a robotic sorting task: the RL agent learns in an extremely simplified simulation and is deployed in the real world without any transfer learning. Using a deep set based encoder allows us to learn the policy with RL based on the outputs of the pre-trained object detector YOLOv3~\citep{redmon2018yolov3}. The policy is executed via the likewise pre-trained GG-CNN~\citep{doug} grasp synthesis network.}
    \label{fig:abstract}
    \vspace{-6mm}
\end{figure}

Moreover, decades of progress in robotics and computer vision have led to many effective solutions to isolated problems. 
Combining these modular systems together to solve a robotics task usually involves significant engineering effort, and the final hand-engineered behaviors can be prone to fail in unforeseen situations \citep{morrison2017cartman}.
Ideally, robots would learn to make use of existing solutions to sub-problems, optimizing their decision making for the task with the tools available. 

To this end, in order to make reinforcement learning more feasible for robotics tasks, we propose an approach relying on a \textit{deep sets} network architecture for decoupling perception and low-level control from the high-level decision-making aspects of a task. 
We demonstrate that while the overall task and reward signal remain the same, the high-level observation and action spaces dramatically reduce the time required to learn a good policy. 

By abstracting away the details that are difficult to simulate and for which high-performing techniques are available, we are able to implement a simulator that recreates the high-level aspects of the task, and trains policies in minutes for deployment directly to the real robot environment. The deep sets architecture allows us to effectively incorporate pre-existing methods that deal with the low-level aspects.
Utilizing state of the art modules for each sub-task, we are able to embody this learned policy on the real robot, with no performance loss due to transfer. 

In spite of the short-comings of deep reinforcement learning, its potential for decision making is well established. Q-Learning, in particular, is provably convergent for fully-observed Markov decision processes in the tabular setting~\citep{watkins1992q}. With the addition of deep neural network function approximation \citep{mnih2015human,tesauro1995temporal} and the application to partially observed domains such as vision-based robotics tasks, these convergence guarantees no longer hold. Furthermore, training large convolutional neural networks purely with reward signals is a challenging optimization problem, with state of the art algorithms still requiring millions of samples~\citep{hessel2017rainbow,espeholt2018impala,mnih2016asynchronous}. In this work, we mitigate this by leveraging the strengths of both reinforcement learning and well-established supervised learning.

First, we outline the approach for modularizing robotics tasks for reinforcement learning. We then demonstrate the importance and effectiveness of combining deep sets with this framework using the real-world robot manipulation task of sorting categorized objects. This task relies on 1) recognizing a range of objects, 2) grasping objects robustly, and 3) making decisions; we abstract the task into these three subproblems, and train an effective policy in simulation, which is then successfully deployed directly onto the robot with no further training. This shows that reinforcement learning, when combined with the state of the art in robotics perception and control, can enable robots to quickly learn to complete useful tasks. 

\section{Related Work}

Recently, end-to-end deep reinforcement learning has been shown to learn complex behaviours in virtual environments \citep{mnih2015human,silver2017mastering,lillicrap2015continuous}. However, these approaches require large amounts of experience and can be unstable due to the complex optimization challenges of training large neural networks on reward signals alone. Such approaches are continually being developed, with improvements including improved sample-efficiency and stability \citep{mnih2016asynchronous,hessel2017rainbow,espeholt2018impala,gu2016q,andrychowicz2017hindsight}. 

The application of deep reinforcement learning to robots in the real world is an emerging research area, with significant potential to provide robots with intelligent decision-making tools \citep{gu2017deep, trpolatent, bruce2018learning} with robotic manipulation being a prominent challenge \citep{kalashnikov2018qt}. Sample-efficiency remains a significant limitation, with these approaches requiring millions of training samples, translating to a restrictive amount of training time and hardware resources \citep{levine2018learning,  pinto2016supersizing}. Simulations provide a platform to train robot policies, which can then be transferred, without the implications of training robots in the real world \citep{rusu2016sim, tan2018sim, zhang2016modular}. Restricted simulation accuracy results in an unavoidable reality gap \citep{rastogisample} when attempting to simulate perception and control at a fine-grained level. Ongoing research in this field is concerned with exploring means to improve robustness to environmental changes to more easily handle the intricacies of the real world, such as domain randomization \citep{tobin2017domain,james2017transferring, OpenAIdex}. Despite impressive progress, these approaches are still limited significantly by the resources required.

Outside of the field of reinforcement learning, state-of-the-art algorithms exist which effectively sidestep several issues faced by end-to-end deep reinforcement learning. For example, in robotic grasping, supervised learning has been used to achieve robust, real time grasp synthesis \citep{doug, dexnet2}. Similarly, perception has also been investigated heavily in supervised learning, with object classification and detection systems achieving high accuracy due to the advent of deep convolutional neural networks \citep{krizhevsky2012imagenet, ren2015faster, redmon2018yolov3, liu2016ssd} and large training datasets \citep{deng2009imagenet, lin2014microsoft}.

\begin{figure}
  \centering
  \input{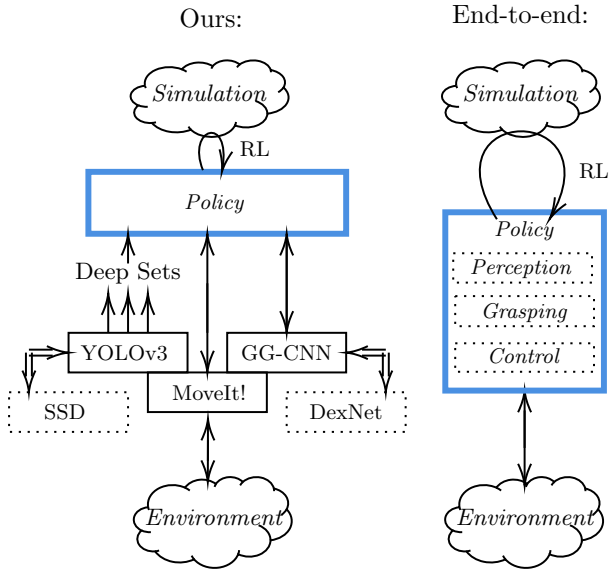}
  \vspace{-1mm}
  \caption{\small Our approach (left) in comparison to standard end-to-end reinforcement learning (right). We propose using deep sets to effectively decouple high-level decision-making from low-level perception and control for which we already have well-performing solutions, to achieve very rapid training and successful deployment to the real world.}
  \label{fig:hero}

\end{figure}

These systems have been shown to be effective in their isolated domains but are task-agnostic, and for them to be functional for real robot tasks, intelligent behaviors such as those provided by reinforcement learning are required. Involving these effective sub-systems in the interface between decision-making and the real world is a focal point of this paper. We propose to make use of deep reinforcement learning for high-level decision-making, leaving low-level perception and control skills to pre-trained and existing off-the-shelf tools (Fig. \ref{fig:hero}). Previous work exists on factoring out the low-level control and perception in robots in order to facilitate reinforcement learning in the real world \citep{lange2012autonomous, kim2004autonomous, gordon2018iqa, hong2018virtual}, although such approaches have only been applied in heavily constrained environments and were limited by the subsystems used during training time. Making use of such predefined modules reflects the \textit{priors} that humans utilize when quickly learning new skills \citep{dubey2018investigating}, and we demonstrate that similar techniques can enable rapid learning in artificial agents as well.

A difficulty that could arise from combining subsystems with neural network policies is that they may provide variable-length observations which are incompatible with the fixed input lengths required by traditional neural networks. A set-based network architecture known as \textit{deep sets} has recently been proposed to handle variable-length and order-invariant inputs such as detections~\citep{DeepSets}. This has been successfully utilized in a simulated game environment to encode ground-truth objects present in the game \citep{woof2018learning}. We propose a novel way to use this technique by encoding outputs from multiple pre-trained perception or control subsystems to be used by deep neural network policies deployed on a real robot. Not only does such a framework eliminate the need for consistent fixed length outputs from the individual subsystems, but it also has the potential to allow subsystems to be added, removed or replaced as required while reusing the \textit{same} existing learned policy.

\section{Approach}

A robot task in the real world may be formulated as a Partially Observable Markov Decision Process (POMDP), with the robot choosing \textit{actions} based on the \textit{observations} it receives from the environment. When the robot acts with the intent of maximizing some \textit{reward function} that indicates progress toward the completion of a task, reinforcement learning methods can be applied to learn the optimal behavior policy. 

We show that with our approach reinforcement learning can be an effective method for learning functional behavior in robotics domains, by making effective use of the many tools that research in robotics and computer vision has generated. In order to make use of these systems and to create a reinforcement learning framework for solving robotics problems, we divide the task into distinct modules: perception, decision-making, planning and control. We would like to be able to utilize the state-of-the-art in each module, and, with the help of a deep sets encoder, combine them using a learned policy trained to maximize reward for a specific task regardless of the tools. Learning this policy enables us to solve decision-making problems that may not be possible with analytic methods, and to quickly solve tasks that would typically take many hours to train using end-to-end learning from scratch. This is in contrast to popular end-to-end approaches, which attempt to learn policies that map directly from raw observations to low-level actions.

\subsection{Vision}

Using vision in robotics is appealing as it an information-rich sensor modality that is commonly available and inexpensive. Furthermore, it holds significant potential for perceiving objects in unstructured real world environments. Utilizing a pre-existing object detection system for perception allows us to extract the important semantic information within the image, without the need for training an end-to-end CNN-based policy. A typical image detection system will take in an image frame, and return a set of proposed detections in the scene, which can be in the form of a \emph{variable} number of bounding boxes \citep{redmon2016you}. Such a module can be trained in a supervised manner, which is significantly more data-efficient than end-to-end RL. This variability poses a challenge for typical deep reinforcement learning policy architectures, which generally have feedforward or convolution filter input layers requiring inputs of a-priori known and fixed size \citep{mnih2015human,mnih2016asynchronous,lillicrap2015continuous,schulman2015trust}.

\subsection{Robot Control}

Robot control is a similarly well-established area of research. In particular, research in the manipulation field has yielded several approaches to robot grasping that could be used off-the-shelf \citep{doug,dexnet2}, as well as effective methods of robot control \citep{sucan2013moveit}. We formulate a set of discrete actions from which a reinforcement learning policy can select, which invoke these control modules, such as grasping an object of a particular class or moving to a pre-defined location.

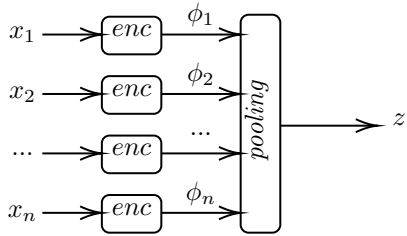
\begin{figure}
  \centering
  \tikzset{every picture/.style={line width=0.75pt}} 

\begin{tikzpicture}[x=0.75pt,y=0.75pt,yscale=-1,xscale=1]

\draw    (130,70) -- (158,70) ;
\draw [shift={(160,70)}, rotate = 180] [color={rgb, 255:red, 0; green, 0; blue, 0 }  ][line width=0.75]    (10.93,-3.29) .. controls (6.95,-1.4) and (3.31,-0.3) .. (0,0) .. controls (3.31,0.3) and (6.95,1.4) .. (10.93,3.29)   ;

\draw    (130,100) -- (158,100) ;
\draw [shift={(160,100)}, rotate = 180] [color={rgb, 255:red, 0; green, 0; blue, 0 }  ][line width=0.75]    (10.93,-3.29) .. controls (6.95,-1.4) and (3.31,-0.3) .. (0,0) .. controls (3.31,0.3) and (6.95,1.4) .. (10.93,3.29)   ;

\draw    (130,130) -- (158,130) ;
\draw [shift={(160,130)}, rotate = 180] [color={rgb, 255:red, 0; green, 0; blue, 0 }  ][line width=0.75]    (10.93,-3.29) .. controls (6.95,-1.4) and (3.31,-0.3) .. (0,0) .. controls (3.31,0.3) and (6.95,1.4) .. (10.93,3.29)   ;

\draw    (130,160) -- (158,160) ;
\draw [shift={(160,160)}, rotate = 180] [color={rgb, 255:red, 0; green, 0; blue, 0 }  ][line width=0.75]    (10.93,-3.29) .. controls (6.95,-1.4) and (3.31,-0.3) .. (0,0) .. controls (3.31,0.3) and (6.95,1.4) .. (10.93,3.29)   ;

\draw   (160,64.8) .. controls (160,62.7) and (161.7,61) .. (163.8,61) -- (186.2,61) .. controls (188.3,61) and (190,62.7) .. (190,64.8) -- (190,76.2) .. controls (190,78.3) and (188.3,80) .. (186.2,80) -- (163.8,80) .. controls (161.7,80) and (160,78.3) .. (160,76.2) -- cycle ;
\draw   (160,154.8) .. controls (160,152.7) and (161.7,151) .. (163.8,151) -- (186.2,151) .. controls (188.3,151) and (190,152.7) .. (190,154.8) -- (190,166.2) .. controls (190,168.3) and (188.3,170) .. (186.2,170) -- (163.8,170) .. controls (161.7,170) and (160,168.3) .. (160,166.2) -- cycle ;
\draw   (160,124.8) .. controls (160,122.7) and (161.7,121) .. (163.8,121) -- (186.2,121) .. controls (188.3,121) and (190,122.7) .. (190,124.8) -- (190,136.2) .. controls (190,138.3) and (188.3,140) .. (186.2,140) -- (163.8,140) .. controls (161.7,140) and (160,138.3) .. (160,136.2) -- cycle ;
\draw   (160,94.8) .. controls (160,92.7) and (161.7,91) .. (163.8,91) -- (186.2,91) .. controls (188.3,91) and (190,92.7) .. (190,94.8) -- (190,106.2) .. controls (190,108.3) and (188.3,110) .. (186.2,110) -- (163.8,110) .. controls (161.7,110) and (160,108.3) .. (160,106.2) -- cycle ;
\draw    (190,70) -- (228,70) ;
\draw [shift={(230,70)}, rotate = 180] [color={rgb, 255:red, 0; green, 0; blue, 0 }  ][line width=0.75]    (10.93,-3.29) .. controls (6.95,-1.4) and (3.31,-0.3) .. (0,0) .. controls (3.31,0.3) and (6.95,1.4) .. (10.93,3.29)   ;

\draw    (190,100) -- (228,100) ;
\draw [shift={(230,100)}, rotate = 180] [color={rgb, 255:red, 0; green, 0; blue, 0 }  ][line width=0.75]    (10.93,-3.29) .. controls (6.95,-1.4) and (3.31,-0.3) .. (0,0) .. controls (3.31,0.3) and (6.95,1.4) .. (10.93,3.29)   ;

\draw    (190,130) -- (228,130) ;
\draw [shift={(230,130)}, rotate = 180] [color={rgb, 255:red, 0; green, 0; blue, 0 }  ][line width=0.75]    (10.93,-3.29) .. controls (6.95,-1.4) and (3.31,-0.3) .. (0,0) .. controls (3.31,0.3) and (6.95,1.4) .. (10.93,3.29)   ;

\draw    (190,160) -- (228,160) ;
\draw [shift={(230,160)}, rotate = 180] [color={rgb, 255:red, 0; green, 0; blue, 0 }  ][line width=0.75]    (10.93,-3.29) .. controls (6.95,-1.4) and (3.31,-0.3) .. (0,0) .. controls (3.31,0.3) and (6.95,1.4) .. (10.93,3.29)   ;

\draw    (250,116) -- (297,116) ;
\draw [shift={(299,116)}, rotate = 180] [color={rgb, 255:red, 0; green, 0; blue, 0 }  ][line width=0.75]    (10.93,-3.29) .. controls (6.95,-1.4) and (3.31,-0.3) .. (0,0) .. controls (3.31,0.3) and (6.95,1.4) .. (10.93,3.29)   ;

\draw   (250,166) .. controls (250,168.21) and (248.21,170) .. (246,170) -- (234,170) .. controls (231.79,170) and (230,168.21) .. (230,166) -- (230,64) .. controls (230,61.79) and (231.79,60) .. (234,60) -- (246,60) .. controls (248.21,60) and (250,61.79) .. (250,64) -- cycle ;

\draw (120,70) node  [align=left] {$\displaystyle x_{1}$};
\draw (120,100) node  [align=left] {$\displaystyle x_{2}$};
\draw (120,160) node  [align=left] {$\displaystyle x_{n}$};
\draw (120,130) node  [align=left] {...};
\draw (175,68) node  [align=left] {$\displaystyle enc$};
\draw (175,158) node  [align=left] {$\displaystyle enc$};
\draw (175,128) node  [align=left] {$\displaystyle enc$};
\draw (175,98) node  [align=left] {$\displaystyle enc$};
\draw (238,113) node [rotate=-270] [align=left] {\textit{pooling}};
\draw (210,60) node  [align=left] {$\displaystyle \phi _{1}$};
\draw (210,90) node  [align=left] {$\displaystyle \phi _{2}$};
\draw (210,150) node  [align=left] {$\displaystyle \phi _{n}$};
\draw (210,120) node  [align=left] {...};
\draw (310,113) node  [align=left] {$\displaystyle z $};

\end{tikzpicture}
\vspace{3mm}
  \caption{\small Deep Sets: The variable number of object detections, $x_i$, are all passed through the same encoder and pooled. The resulting fixed-sized representation vector, $z$, is concatenated with additional state information and passed to the policy as the observation}
  \label{fig:ds}

\end{figure}

\subsection{Modular Reinforcement Learning Framework}

Our framework combines the perception and control modules using a novel set-based reinforcement learning approach. The input image is fed into the vision module, resulting in a variable number of possible object detections.
\subsubsection*{\textbf{Deep Sets}} In order to handle this variable-sized output, we utilize a \textit{deep sets} encoder to learn a representation of the detection instances that can then be used with a simple feedforward network \citep{DeepSets}. This representation is computed by passing class labels through a shared encoder, followed by a permutation-equivariant pooling operation, such as sum, mean or max pooling, over all encoded instances (Fig. \ref{fig:ds}). This set encoding is combined with any additional state information, such as robot position, before being used as the input to a neural network policy. The policy can then select an action that can make use of the robot control module.

This abstracted decision-making task allows us to train the policy very quickly in a simplified simulated environment. After training, the policy can be then be transferred directly to the real robot with no further training required, without significant loss in performance.

The proposed framework is not specific to a particular problem and can be applied to any domain in which perceptual and motor priors are available. In this paper, we evaluate our framework on a robot sorting task.

\section{Experimental Implementation}

To evaluate this modular approach, we consider an object sorting task that relies on robust object detection, grasping and decision-making. The setup consisted of a large bin, divided into three equally sized regions depicting the fixed number of classes. At the beginning of a run, three classes of objects, corresponding to the bin locations, were dispersed randomly in the bin. An episode was considered complete when all objects were in their assigned bins. The robot moved between three set positions above each region, with an eye-in-hand RGB-D camera. State-of-the art methods were used for the off-the-shelf components of our architecture. Rewards were provided automatically by the environment on a per-action basis, with a reward of +1 provided for grasping or dropping an object appropriately, and -1 for grasping or dropping an incorrect object.

Two types of perception modules were used for this experiment. The first consisted of a simple color-based object classifier. This was used to detect a variety of objects with adversarial geometry mentioned in \citep{dexnet2}. These objects were designed to stress test grasping systems, and were chosen for this task to reflect the difficulty of grasping real world objects. The second consisted of a pre-trained YOLOv3 network \citep{redmon2018yolov3}, a fast and accurate image detection network, trained on the Common Objects in Context (COCO) dataset \citep{lin2014microsoft}. This dataset contains the three categories of everyday objects used in our experiments: \textit{Teddy Bears, Balls and Bottles}. It runs at over 25 frames per second, allowing our system to receive the bounding boxes and class labels of the variable number objects within the camera frame in real time, when invoked.

\begin{figure}
    \centering
    \begin{subfigure}[b]{0.22\textwidth}
        \centering
        \includegraphics[height=\textwidth]{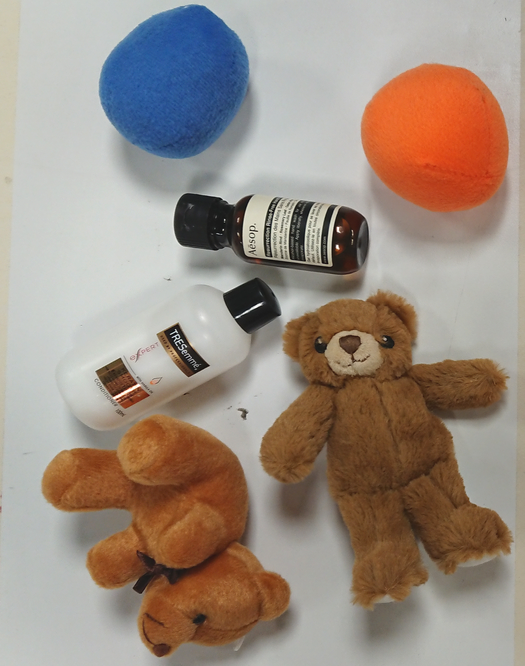}
        \caption{\small Real world objects detected with YOLOv3}
        \label{sub:real}
    \end{subfigure} 
    \begin{subfigure}[b]{0.22\textwidth}
        \centering
        \includegraphics[height=\textwidth]{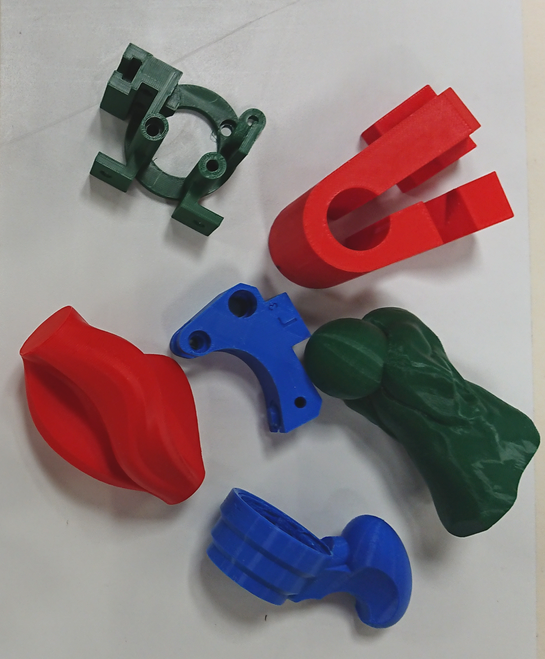}
        \caption{Adversarial objects detected with color thresholds}
        \label{sub:adv}
    \end{subfigure}
    \label{fig:realworld}
    \caption{\small Real world experimental objects}
\end{figure}

For the grasping module, depth images were used for producing grasp proposals using the GG-CNN grasping network \citep{doug}. GG-CNN is a real-time grasping system which outputs a heat map where points correspond to grasp quality. The highest ranking grasp proposal that falls within a bounding box is invoked by the policy if an object of that class label is selected to be grasped. 

Based on the perception and grasping information given, the agent then chose one of 7 possible actions: move to each of the 3 predefined robot positions, grasp one of the 3 possible object classes, or drop a currently grasped object. If a chosen action was infeasible (such as a drop action being chosen when there was no object grasped), no action was performed and no reward was awarded.

\begin{figure}
  \centering
  \includegraphics[width=\columnwidth]{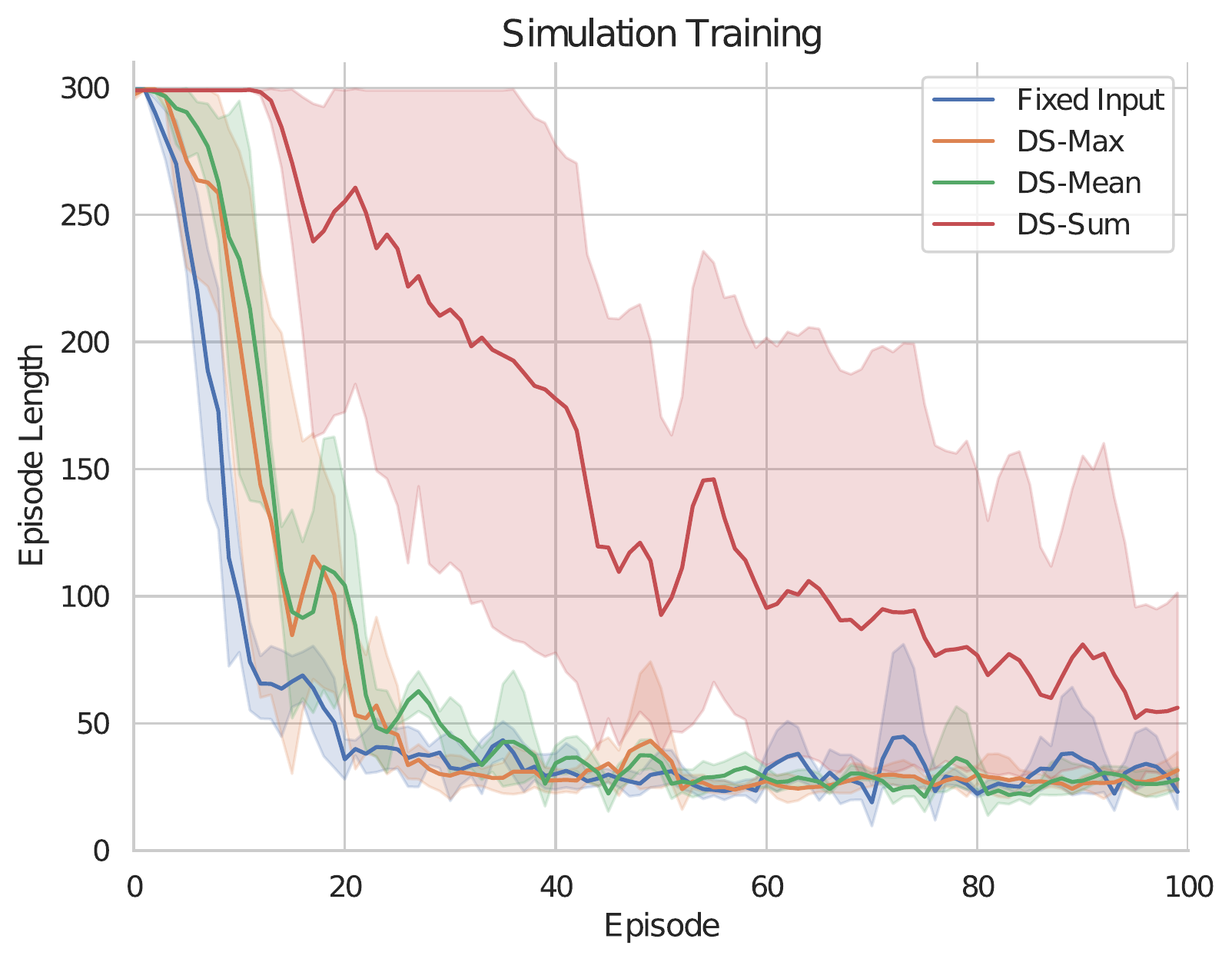}

  \caption{\small Time to Solve for different input types}
  \label{fig:learning}
\end{figure}

\subsection{Simulation Setup}
A simulation of the experimental setup was designed for training the reinforcement module, with the perception and control modules abstracted out. Instead, the agent was given the ground truth object set and perfect action execution was assumed. Fig. \ref{fig:abstract} depicts a rendering of the task simulation. The different object categories were represented by colored circles for visual interpretation only; the agent itself merely perceived the class labels, with no direct meaning assigned to the objects. Reward is easily retrieved from the environment, as the simulator has full state information. This type of setup, with direct abstracted information as input, facilitated efficient policy learning, resulting in merely 10 minutes of training time.

\subsection{Real World Setup}
For evaluating the trained agent in the real world, we used a 7DOF Franka Emika Panda robot arm, equipped with an Intel Realsense D435 depth camera. An open area with three distinct `bins' was designated within the robot's workspace. 
Objects were shuffled within the bins at the start of an experiment, and the robot began in the centre bin. In this case, the rewards were determined based on the vision sub-system, which was used to identify object classes. This setup is not dependent the specific arm used, the same setup was tested on a Kinova Mico arm with similar results.

\begin{figure*}[h]
    \centering
    \begin{subfigure}[b]{0.3\textwidth}
        \includegraphics[width=1\textwidth]{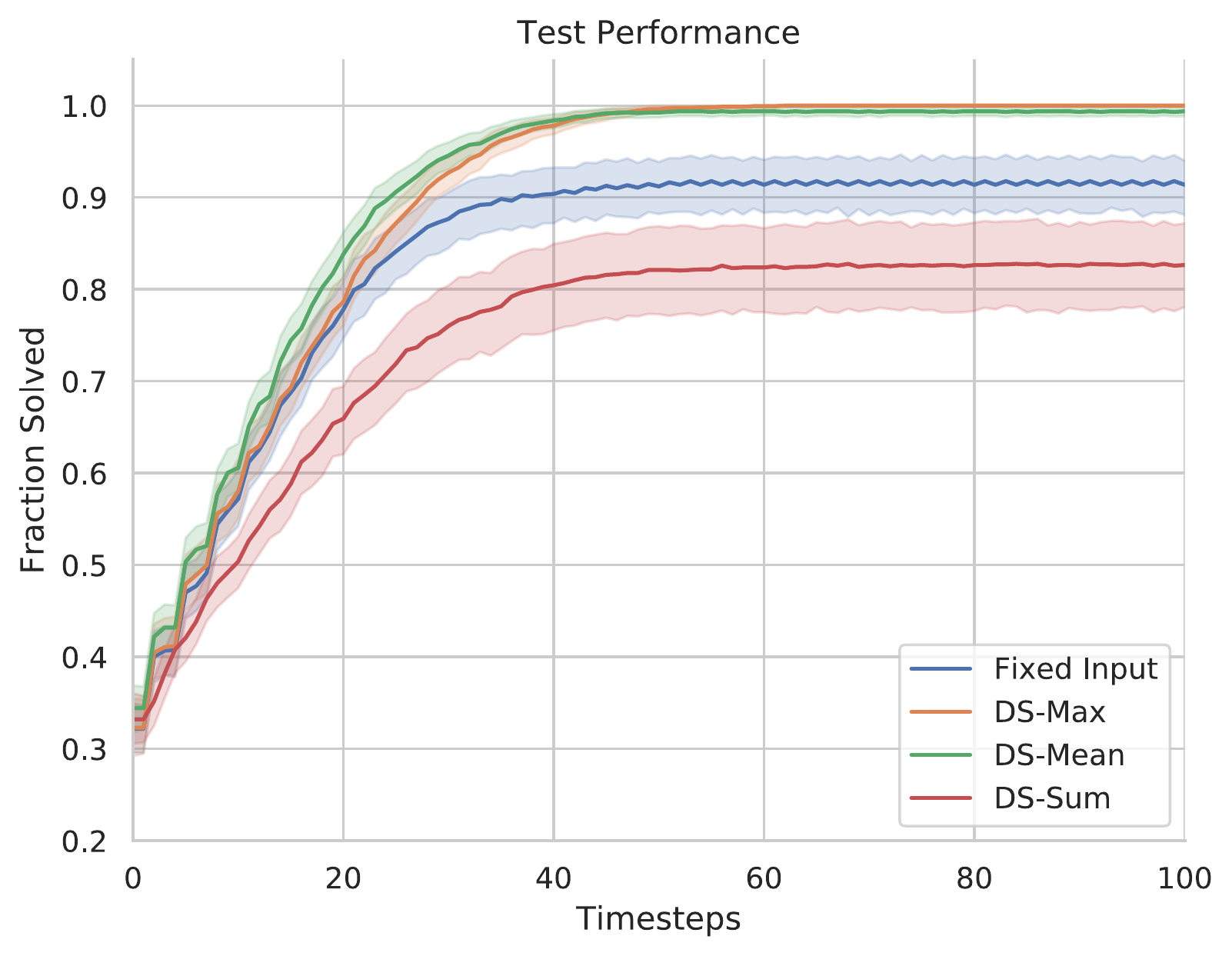}
        \caption{\small 5 Objects}
        \label{fig:5}
    \end{subfigure}
    ~ 
    ~ 
    \begin{subfigure}[b]{0.3\textwidth}
        \includegraphics[width=\textwidth]{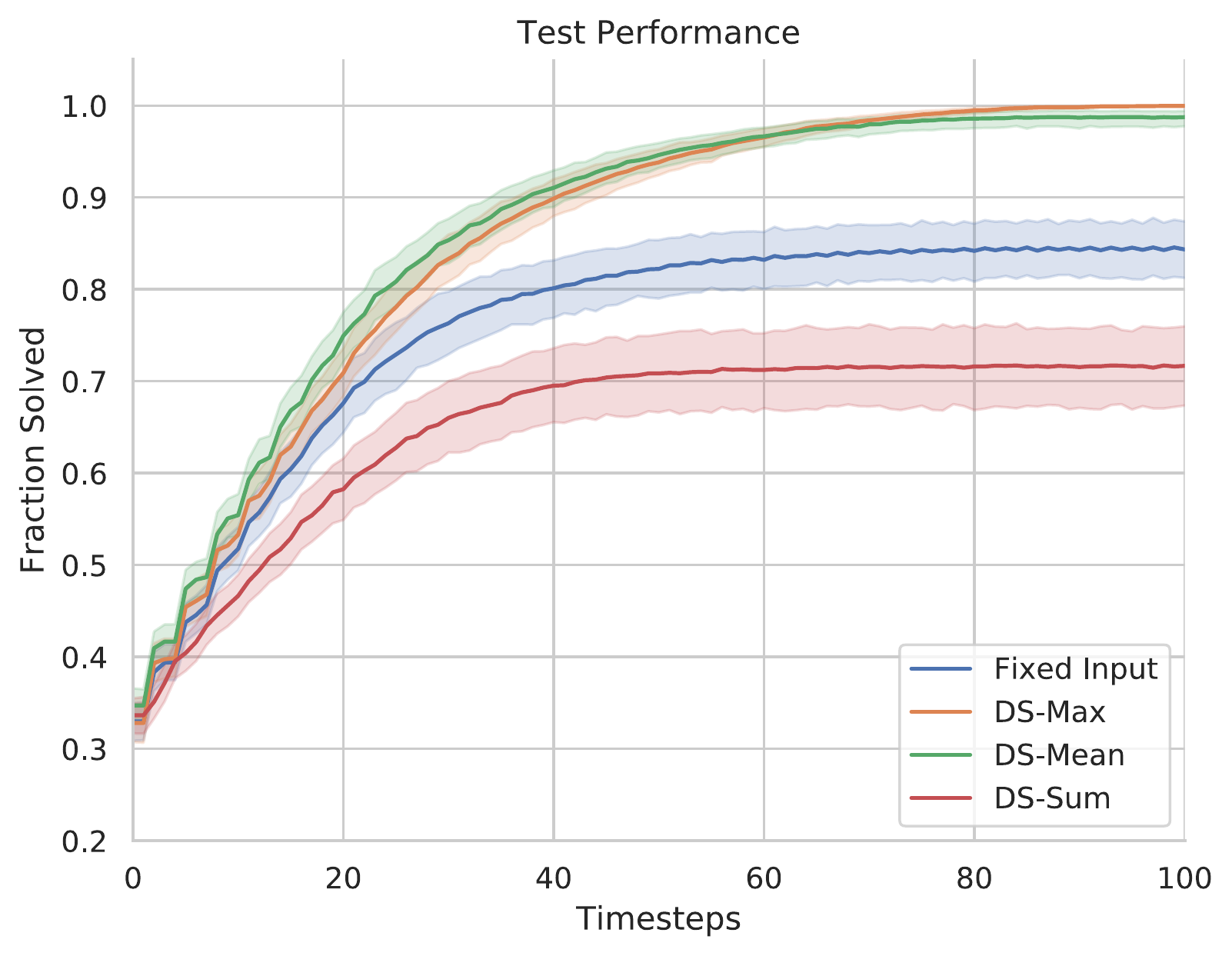}
        \caption{\small 10 Objects}
        \label{fig:10}
    \end{subfigure}   
    ~ 
    \begin{subfigure}[b]{0.3\textwidth}
        \includegraphics[width=\textwidth]{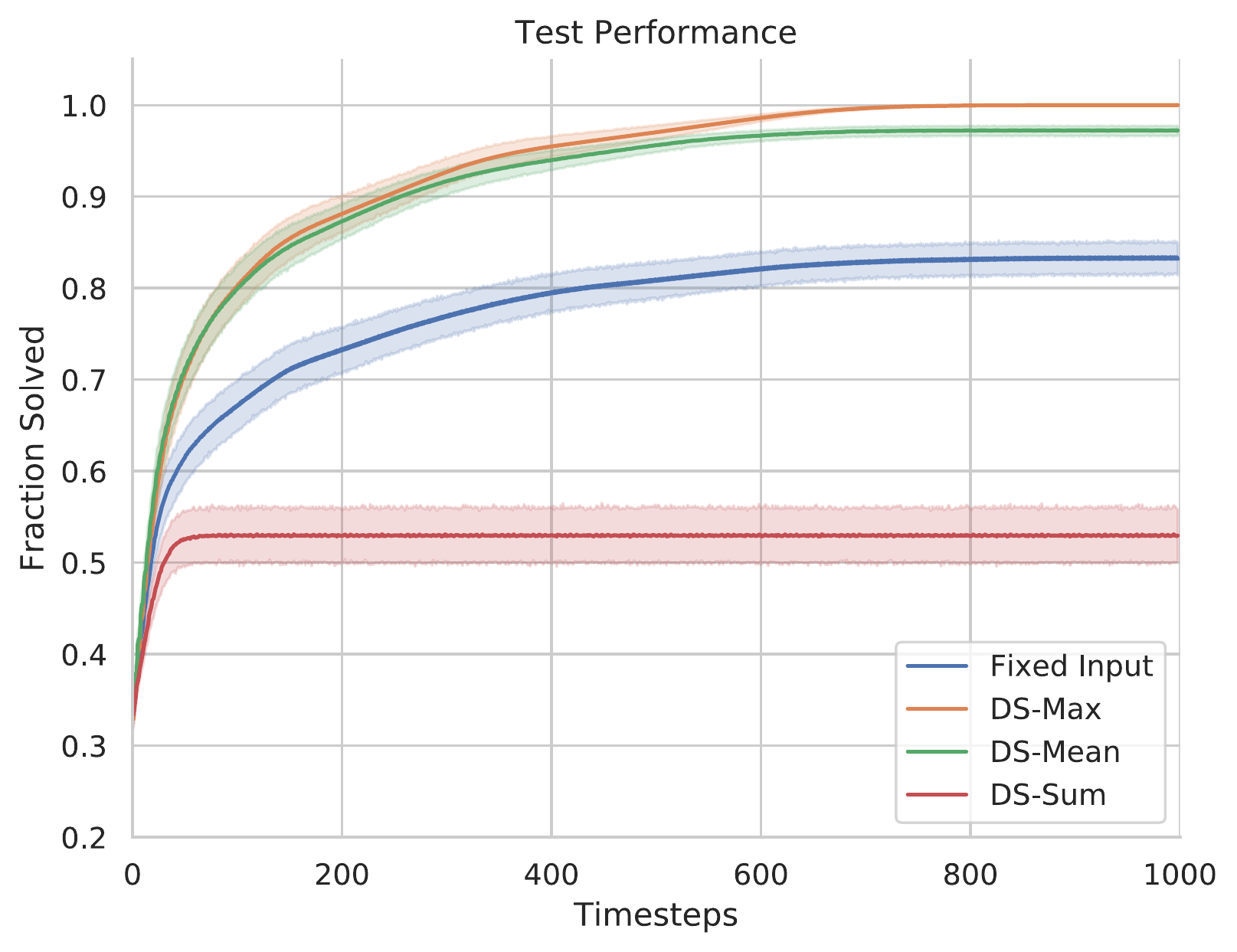}
        \caption{\small 100 Objects}
        \label{fig:100}
    \end{subfigure}
    \caption{Generalization of Different Policies}\label{fig:generalization}
\end{figure*}

\subsection{Agent Architecture}
For our experiments, we used DQN \citep{mnih2015human} as the base reinforcement learning algorithm.
The agent consists of a fully-connected instance encoder network of 2 layers, followed by a pooling operation \citep{DeepSets}. This encoding is concatenated with the agent state and passed through another encoding layer before finally being input to a 2 layer fully connected Q-Network. As the agent can only perceive a single bin at a time, a frame stacking of 4 is used for the agent's observations, as in \citep{mnih2015human}. These stacked observations are passed as input to the final encoding layer before the Q-Network. 

The agent was trained for 100 episodes over 5 random seeds, with each episode terminating after a maximum of 300 timesteps.
Due to the short training period, there is no limit placed on replay buffer capacity, as the training steps are significantly less than required in \citep{mnih2015human}. A comprehensive list of hyperparameters used is given in Table \ref{tab:hypers}. Excluding the choice of pooling method however, we found that overall agent performance was insensitive to the choice of precise hyperparameter values.

Overall learning time corresponded to approximately 10 minutes in the real world. Training was performed on the CPU of a commercially available desktop computer; further speedups may be possible by use of dedicated hardware such as GPUs, but we did not find this to be necessary.

\begin{table}[]
\centering
\caption{List of hyperparameters}
\label{tab:hypers}
\begin{tabular}{@{}llll@{}} 
\toprule
\textbf{Hyperparameter} & \textbf{Value} \\ 
\midrule
Episode time limit &  300 \\
Instance embedding size      &  128 \\
State embedding size & 128 \\
Discount factor & 0.9 \\
Replay memory size & $\infty$ \\
Frame stacking & 4 \\
Mini-batch size & 64 \\
Learning rate & 0.0001 \\
Initial $\epsilon$ exploration & 1.0 \\
Final $\epsilon$ exploration & 0.05 \\
Maximum episodes $\epsilon$ is annealed over & 20 \\
Maximum training episodes & 100 \\
\bottomrule                    
\end{tabular}
\end{table}

\section{Results}

We evaluated the efficacy of our proposed architecture in terms of the time taken to train to convergence and the generalization capabilities across variable numbers of object instances.

\subsection{Training: Comparison of Pooling Types}

In order to assess the effectiveness of \textit{deep sets} as a method for handling the variable number of object detections that an image detection system may provide, we compare a \textit{deep sets} agent to a baseline agent: a feedforward network with a fixed input size. The pooling operation used for the \textit{deep sets} architecture depends on the type of task \citep{DeepSets}. Therefore, to determine the optimal pooling type for our case, we compare sum, mean and max pooling.

The pooling methods can be defined as:

\begin{equation*}
\begin{aligned}
\textit{Max pooling} &: \max\limits_{j}(\phi_{i,j}) \\
\textit{Mean pooling} &: \frac{\sum_{j=1}^{N_{\text{obs}}}\phi_{i,j}}{N_{\text{obs}}} \\
\textit{Sum pooling} &: \sum_{j=1}^{N_{\text{obs}}}\phi_{i,j}
\end{aligned}
\end{equation*}

where $\phi_{j}$ is the feature encoding of the $j$th object instance, $i$ is the feature dimension and $N_{\text{obs}}$ is the total number of observed instances.

Each policy, originally trained with three objects per bin (nine objects in total), is then deployed in simulation environments with 5, 10 and 100 objects per bin. The input size of the baseline method was set to the maximum number of objects that the policy would encounter (a total of 300).

The baseline method converges to a near optimal policy at slightly faster rate when compared to the other methods, shown in Fig. \ref{fig:learning}. However, as seen in Fig. \ref{fig:generalization}, its performance when deployed decreases significantly with increase in the number of objects per bin. This is expected, as the network input is representative of all observed instances, and is not required to first learn an adequate representation of the pooled instances, contrary to deep sets. With a maximum of nine objects observed at once during training, the remaining input neurons are never activated, and the corresponding weights are never updated. Furthermore, this approach is sensitive to permutations, with potential overfitting to the specific orders and combinations of objects seen during training. 

On the other hand, the deep sets agents are shown to be better equipped to handle the increasing number of objects. In particular, max pooling performance does not appear to deteriorate at all. For this task, it should not matter to the policy how many instances of an object class are observed, and max pooling conveys the presence of relevant features, rather than their quantities. Mean pooling is able to encode both the presence and the ratios of the different observed instances, and thus performs well. However, the increasing magnitude of object occurrences, leads to mean values that are scaled differently than during training. Sum pooling performs the worst, due to high sensitivity to out-of-distribution magnitudes of pooled instance embeddings.

The inherent encoding capabilities of the three pooling types are reflected in the training times: sum pooling trains the slowest as it has to adapt to the variance in the unconstrained input representations, whereas max and mean pooling learn quickly due to their representations being less sensitive to input sizes.

\begin{figure}
  \centering
  \includegraphics[width=\columnwidth]{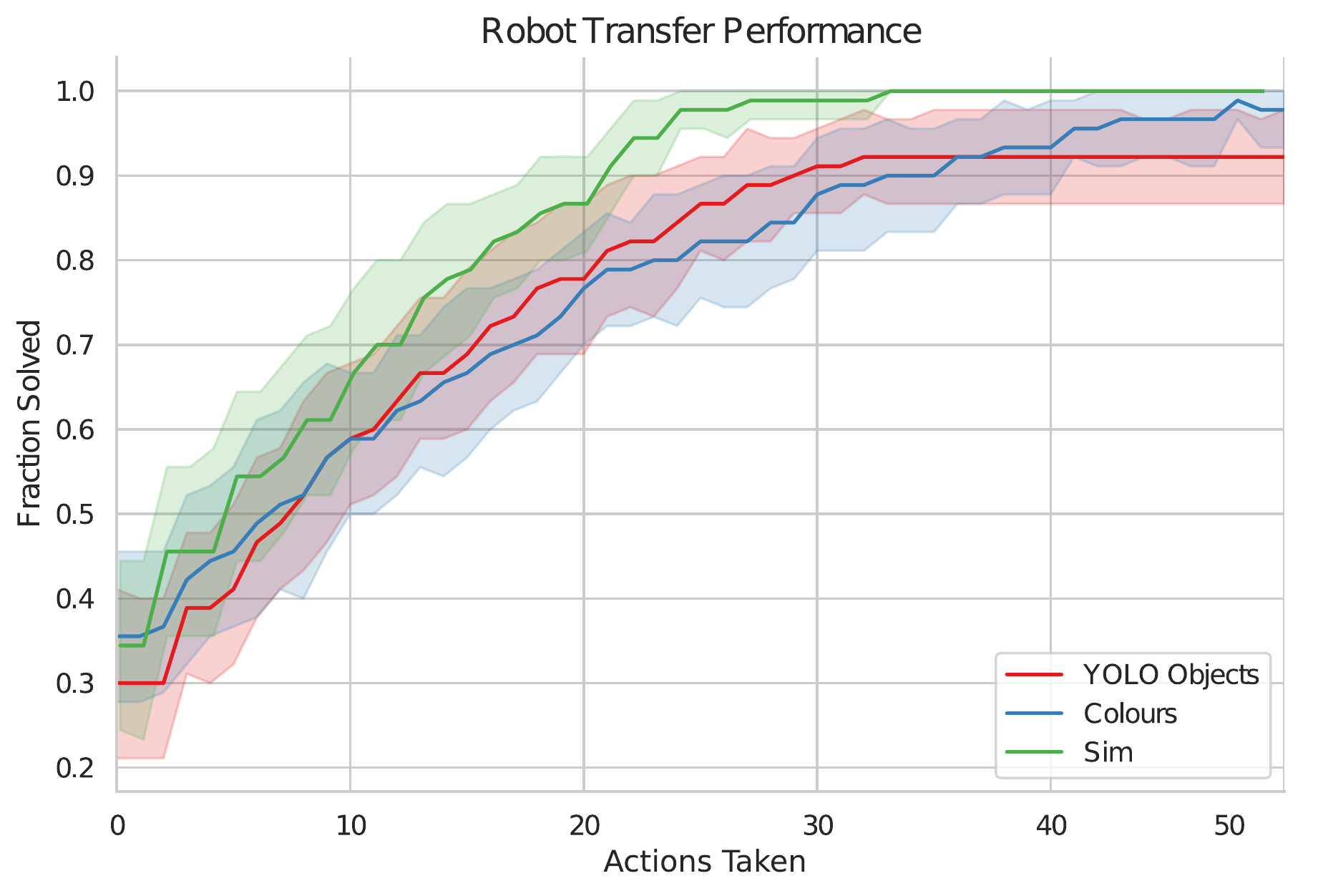}

  \caption{\small Time to Solve for different environments}
  \label{fig:robottest}
\end{figure}

\subsection{Transfer to Real World}

After training in simulation for 100 episodes, we demonstrate that the learned policy can be transferred directly to the real robot environment (Fig. \ref{fig:robottest}). We show performance by plotting the fraction of objects in the correct location for each action in the episode. The performance of the same policy, when embodied with two different perception systems for their corresponding object sets, is shown alongside the average simulation performance. Direct deployment of the trained policy on the real world tasks yielded performance comparable to that in simulation in spite of the challenges of the unstructured environment. 

\begin{figure}
    \centering
    \begin{subfigure}[b]{0.15\textwidth}
        \includegraphics[height=\textwidth]{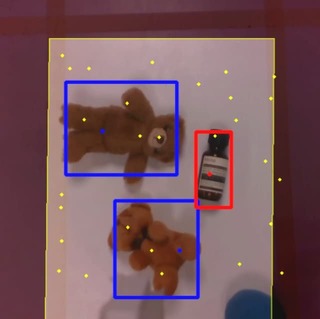}
        \caption{\small Perception}
        \label{fig:p}
    \end{subfigure}
    \begin{subfigure}[b]{0.15\textwidth}
        \includegraphics[height=\textwidth]{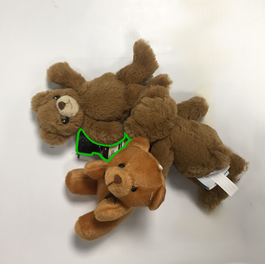}
        \caption{\small Occlusion}
        \label{fig:o}
    \end{subfigure}   
    \begin{subfigure}[b]{0.15\textwidth}
        \includegraphics[height=\textwidth]{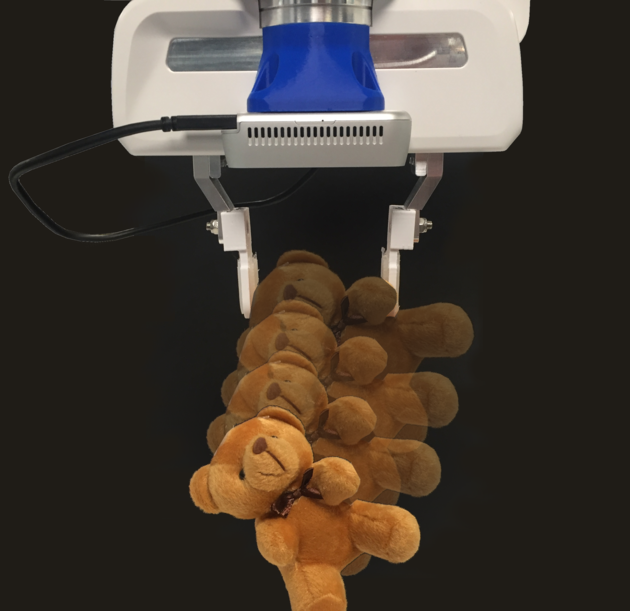}
        \caption{\small Grasp}
        \label{fig:d}
    \end{subfigure}
    \caption{\small Potential failure modes. \textbf{(a)} Objects and grasp detections are imperfect, the blue ball on the right has been missed, \textbf{(b)} objects may stack causing occlusion and  \textbf{(c)} grasps may fail}\label{fig:failmodes}

\end{figure}

In the YOLO experiment, the objects being sorted can be seen in Fig. \ref{sub:real}. The discrepancies between the simulated performance and the YOLO performance can be attributed to perception failures such as the vision system not recognizing an object (Fig. \ref{fig:p}) or occlusions (Fig. \ref{fig:o}). The Color experiment involved adversarial objects, designed to robustly test the performance of grasping systems (Fig. \ref{sub:adv}). The discrepancies in this case can largely be attributed to failed grasps (Fig. \ref{fig:d}).

These failure cases are at this stage unavoidable in the real world, however this method has yielded impressive robot transfer. Another benefit of this framework is the modular components which can be easily switched out as the state of the art modules improve. In the experiments, color thresholding to find bounding boxes for the adversarial objects was seamlessly interchanged with detecting real world objects with YOLOv3 \citep{redmon2018yolov3}. As object detectors improve, bounding boxes can be suggested by improved systems, or different objects can be substituted in, with no retraining of the policy. Similarly, as robotic grasping improves the module can be swapped out with no retraining. 

\section{Conclusion}

In this work, we propose real world robot solutions with a modular network that is able to effectively handle variable-length inputs, allowing the use of pre-existing low-level modules in combination with reinforcement learning. By allowing reinforcement learning to take care of the decision making and utilizing pre-trained or fully determinable modules, the agent is no longer required to learn complex associations between raw input and low-level behaviour. By decoupling the perception and control modules from the real world to simulation and using a deep set method to encode the input, the reinforcement learning algorithm can be quickly and efficiently trained in simulation and the policy directly implemented onto a robot. As the decoupled modules can be trained in a supervised way or determined mathematically, the execution is an effective, high performing, real world robotic solution. We utilize deep sets to process the variable number of observations the agent may encounter, and train the agent to convergence in simulation in only minutes. We show that when embodied on a real robot with effective perception and grasping modules, the agent can be deployed directly with no further training, and only a slight decrease in performance due to the complexities of real world environments.

In order to improve the performance of our framework, the subsystems could be fine-tuned during deployment time. This will likely still be faster than learning end-to-end as it would require fewer data samples to refine an existing trained network \citep{rusu2016sim}.

Another approach could be to employ multiple subsystems for perception to increase redundancy and therefore decrease chances of perception failures. This is akin to using multiple modalities to improve robustness of vision systems \citep{eitel2015multimodal}.

Active exploration may be used to see around occlusions. This could either be in the form of incorporating a pre-existing sub-system with exploration capabilities, or by learning exploration on the reinforcement learning side of the framework. 

Finally, we would also like to implement our framework on other robotic domains such as navigation to evaluate its generalization capabilities across robotic domains. 

\newpage
\bibliographystyle{named}
\bibliography{example.bib}

\vspace{12pt}

\end{document}